\documentclass[11pt,a4paper]{article}
\usepackage[hyperref]{naaclhlt2019}
\usepackage{times}
\usepackage{latexsym}

\usepackage{amssymb}
\usepackage{multirow}
\usepackage{url}
\usepackage{graphicx}

\usepackage{booktabs}
\usepackage{amsmath}
\usepackage{xcolor}
\usepackage[rightcaption]{sidecap}
\usepackage{wrapfig}

\aclfinalcopy 

\title{Multilingual NMT with a language-independent attention bridge}

\author{Ra\'{u}l V\'{a}zquez \quad Alessandro Raganato \quad J\"{o}rg Tiedemann \quad Mathias Creutz \\
  University of Helsinki \\
  Department of Digital Humanities \\
  {\tt \{name.surname\}@helsinki.fi}
  }

\date{}

\begin{document}
\maketitle
\begin{abstract}
In this paper, we propose a multilingual encoder-decoder architecture capable of obtaining multilingual sentence representations by means of incorporating an intermediate {\em attention bridge} that is shared across all languages. That is, we train the model with language-specific encoders and decoders that are connected via self-attention with a shared layer that we call attention bridge. This layer exploits the semantics from each language for performing translation and develops into a language-independent meaning representation that can efficiently be used for transfer learning. We present a new framework for the efficient development of multilingual NMT using this model and scheduled training. We have tested the approach in a systematic way with a multi-parallel data set. We show that the model achieves substantial improvements over strong bilingual models and that it also works well for zero-shot translation, which demonstrates its ability of abstraction and transfer learning.
\end{abstract}

\section{Introduction}

Neural Machine Translation (NMT) provides a powerful approach to MT that achieves high translation accuracy and fluency due to its ability to capture long-distance dependencies and internal abstractions. 
Multilingual Machine Translation addresses the task of building a system that can translate between multiple languages, either by using a \textit{many-to-one} approach with several source languages to translate into a single target language; a \textit{one-to-many}, with a single source language to translate into different target languages; or \textit{many-to-many} models that allow  multiple languages on both sides (see e.g. \citet{dong_multitask-learning_2015}, \citet{zoph_multi-source_2016}, \citet{schwenk_multiling-sentence-repr_2017}). 

NMT certainly provides an ideal setting for multilingual MT because it can efficiently share the parameters of the model and take advantage of the various similarities found by the model in the hidden layers and embeddings \cite{firat_multi-way_2016, johnson_google-zeroshot_2016, blackwood_task-spec-attention_2018}. Besides, multilingual NMT has the advantage of considerably improving the performance of neural translations systems for low-resourced languages \cite{lakew_improving-zero-shot_2017} and it provides the possibility of translating between language pairs that were not seen during training \cite{firat_zero-resource_2016}, commonly called zero-shot translation \cite{johnson_google-zeroshot_2016}. 

NMT as first proposed with an encoder-decoder architecture \cite{sutskever_seq2seq_2014} allows for learning fixed-size sentence representations embedded in continuous vector spaces. Such representations are useful since they provide means for testing downstream tasks, also enabling a deeper linguistic analysis and understanding of what the neural models are learning \cite{conneau_cram-into-1-vector_2018,senteval,3rd-workshop-representation_2018}. However, as those basic models have strong limitations and 
NMT advanced in complexity becoming the de-facto standard in machine translation, the aforementioned sentence representations were replaced by the use of attention mechanisms, i.e., context vectors attending to different parts of the input sentence while generating the output sentence \cite{bahdanau_attention_2014}, and self-attention that replaces recurrent layers in the encoder and decoder \cite{vaswani_attention-all-you-need_2017}.
Nevertheless, it is also possible to create a fixed-size vector representation while still making use of the advantages of attention mechanisms by including a compound attention layer between encoder and decoder \cite{cifka_bleu-VS-meaningRep_2018}. This is an architecture that we adapt in our approach for a multilingual setting of translation.

In this paper we focus on models that allow the translation between many languages, where we outline the development of a 
language-independent representation
based on an {\em attention bridge} that is shared across all languages.
This is in contrast with previous attempts to obtain such a "neural interlingua" \cite{lu_neural-interlingua_2018}, where the authors have only tested theirs under a one-to-many and many-to-one scenario. 
In order to do this, we propose an architecture based on shared self-attention for multilingual NMT with language-specific encoders and decoders, that achieves comparable results to the current state-of-the-art architectures 
and can as well address the task of obtaining language-independent sentence embeddings.
Those embeddings are created from the encoder's self-attention and connect to the language-specific decoders that attend to them, hence the name bridge. We also add a penalty term to avoid redundancy in the shared layer. More details of the architecture are given in section \ref{sec:model_architecture}.

\section{Related Work}
\label{sec:related_work}

Multilingual NMT has been widely studied and  developed in different pathways during the last years \cite{luong_multitask-seq2seq_2015, dong_multitask-learning_2015, chen_teacher-student_2017, johnson_google-zeroshot_2016}. Work has been done with networks that use language specific encoders and decoders, such as \citet{dong_multitask-learning_2015}, who used a separate attention mechanism for each decoder on one-to-many translation. \citet{zoph_multi-source_2016} exploited a multi-way parallel corpus in a many-to-one multilingual scenario, while \citet{firat_multi-way_2016} used language-specific encoders and decoders that share a traditional attention mechansim in a many-to-many scheme. Another approach is the use of universal encoder-decoder networks that share embedding spaces to improve the performance of the model, like the one proposed by \citet{gu_universal_2018} for improving translation on low-resourced languages and the one from \citet{johnson_google-zeroshot_2016}, where the term \textit{zero-shot translation} was coined.

Sentence meaning representation has as well been vastly studied under NMT settings. When introducing the encoder-decoder architectures for MT, \citet{sutskever_seq2seq_2014} showed that the seq2seq models are better at encoding the meaning of sentences into vector spaces than the bag-of-words model. Recent work includes that of  \citet{schwenk_multiling-sentence-repr_2017}, who use multiple encoders and decoders that are connected through a shared layer, albeit with a different purpose than performing translation. In \citet{platanios_contextual_2018} the authors show an intermediate representation that can be decoded to any target language while describing a parameter generation method for universal NMT.  \citet{cifka_bleu-VS-meaningRep_2018} introduced an architecture with a self-attentive layer to extract sentence meaning representations of fixed size. Here we use a similar architecture in a multilingual setting. 

Our work on multilingual MT and sentence representations is closely related to the recently published paper by \citet{lu_neural-interlingua_2018}. There, the authors attempt to build a neural interlingua by using language independent encoders and decoders which share an attentive long short-term memory (LSTM) layer. Our approach differs  because our model  is  able  to  encode  any  sequence  with  variable  length  into  a  fixed  size  representation, without suffering from long-term dependency problems \cite{lin_self-attentive_2017}	 and without the need of padding for downstream task testing. Additionally, we also experiment in a multilingual many-to-many setting, instead of only one-to-many or many-to-one.

\section{Model Architecture}
\label{sec:model_architecture}
In this section we introduce the proposed architecture. Given that we apply some simple modifications to the traditional attention mechanisms \cite{bahdanau_attention_2014}, we will first start by introducing it in its original formulation. After that, we proceed to introduce our architecture by building upon this theory.

In the following, it should be noted that the architecture is not restricted to RNN-based encoders and hence one could make use of CNN- or Transformer-based encoders. We made this choice for the sake of clarity in the formulation.

\subsection{Background: Attention Mechanism}
Given an input $X = (x_1, \hdots ,x_n)$, a sequence of embedded tokens into the vector space $\mathbb{R}^{d_x}$ , our goal is to generate a translation $Y = (y_1, \hdots,y_m)$. 
The \textbf{encoder} is a recurrent neural network (RNN) that sequentially reads each element in $X$ to generate a context vector $c$. Generally, for each token the RNN generates a hidden state $h_t \in \mathbb{R}^{d_h}$ where the last hidden state of the RNN often defines $c$:
\begin{align}
    h_t &= f(x_t, h_{t-1}) \label{eq:encoder_states} \\
      c &= h_n \label{eq:c}
\end{align}
and $f:\mathbb{R}^{d_x} \times \mathbb{R}^{d_h} \longrightarrow \mathbb{R}^{d_h}$ is a non-linear activation function. We use bidirectional LSTM units \cite{hochreiter_lstm_1997} as $f$ in this paper.

Then, the \textbf{decoder} network sequentially computes $(y_1, \hdots,y_m)$ by optimizing
\begin{align}
    p(Y | X ) = \prod_{t=1}^m p(y_t | c, Y_{t-1}) \label{eq:loss}
\end{align}
where $Y_{t-1} = (y_1, \hdots, y_{t-1})$. Each distribution 
$p_t = p(y_t | c, ) \in \mathbb{R}^{d_{v}}$ is usually computed with a softmax function over all the words in the vocabulary, taking into account the current hidden state of the decoder
\begin{align}
    p_t &= softmax(y_{t-1},s_t) \\
    s_t &= \varphi(c, y_{t-1}, h_{t-1}) 
\end{align}
where $\varphi$ is another non-linear activation function and $d_{v}$ is the size of the vocabulary.

Including an \textbf{attention mechanism} in the decoder implies that a different context vector $c_t$ will be computed at each step $t$, instead of fixing $c$ as in equation \eqref{eq:c} for generating all output words. This alignment method allows the decoder to assign different weights to each part of the input at every decoding step \cite{bahdanau_attention_2014} by defining $c_t$ as the weighted sum of hidden states of the encoder $c_t = \sum_{i=1}^n \alpha_{t,i}h_t$, where $\alpha_{t,i}$  indicates how much the $i$-th input word contributes to generating the $t$-th output word, and is usually defined as
\begin{align}
    \alpha_{t,i} &= \frac{exp(e_{t,i})}{\sum_{k=1}^{n}exp(e_{t,k})} \label{eq:alphas} \\
    e_{t,i} &= g(s_t, h_i)  \label{eq:energy}
\end{align}
and $g$ is a feedforward neural network.

\subsection{Multilingual Model}
The multilingual model we propose is a simple extension of the attention-based model previously described, with three major modifications. Namely, \textit{(i)} the incorporation of a self-attention layer (attention bridge), shared among all language pairs, that serves as a neural interlingua; \textit{(ii)} the use of language-specific encoders and decoders for each language pair, trainable with a language-rotating scheduler;  and \textit{(iii)} the introduction of a penalty term to avoid redundancy in the attention heads. We now formally develop on these features.

\textit{(i) Attention bridge:}
The attention bridge must serve as an intermediate layer that encodes, as much as possible, language-independent sentence representations. For this we use the concept of self-attention that has recently been applied in similar ways \cite{lin_self-attentive_2017, vaswani_attention-all-you-need_2017, cifka_bleu-VS-meaningRep_2018, tao_utterance_2018}. For simplicity, let us assume that we have computed some encoder states as in equation \eqref{eq:encoder_states} so that we have a matrix
\begin{align}
    H = ( h_1 , h_2, \hdots, h_n) ~ \in ~ \mathbb{R}^{ d_h \times n }
\end{align}
Similar to \citet{lin_self-attentive_2017}, we encode this variable length sentence-embedding matrix $H$ into a fixed size $M \in \mathbb{R}^{ d_h \times r }$ capable of focusing on $r$ different components of the sentence, defined as follows:
\begin{align}
    B &= softmax\left(W_{2} \mbox{ReLU}(W_{1}H ) \right) \label{eq:B}\\
    M &= BH^T \label{eq:attention_bridge}
\end{align}
where $W_{2} \in \mathbb{R}^{ r \times d_w }$ and $W_{1} \in \mathbb{R}^{ d_w \times d_h}$ are weight matrices, $r$ is the number of attention heads (column vectors) in the attention bridge (matrix $M$), $d_w$ the dimension of the linear transformations needed to compute $B$. Notice that the attention bridge matrix $M \in \mathbb{R}^{ d_h \times r }$ has a fixed size that does not depend on the length of the input sentence $n$. 

\begin{figure}[ht]
\includegraphics[width= 0.9\hsize]{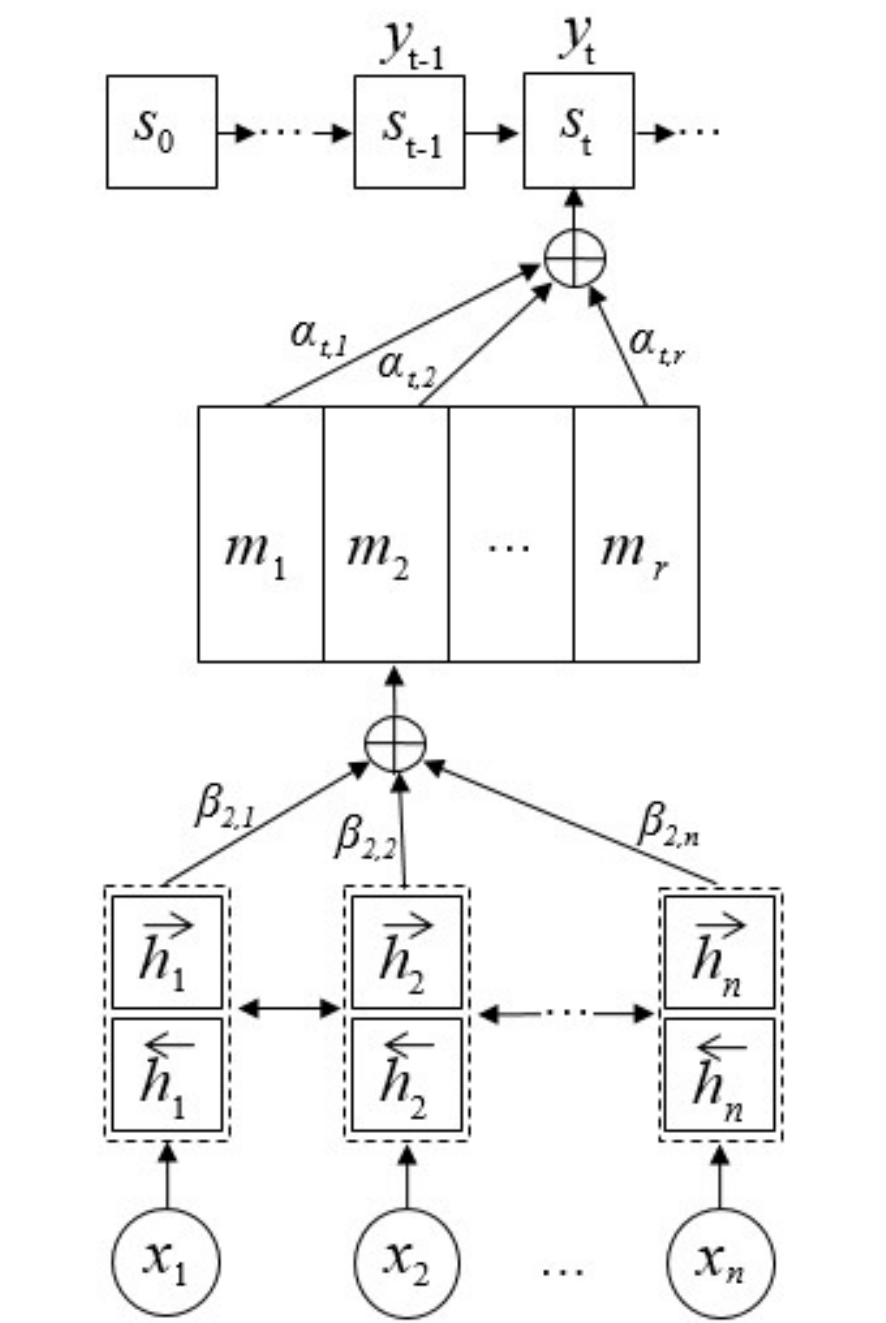} 
\caption{ An overview of the proposed model for one language pair, generating the $t$-th target word $y_t$ given a source sentence $(x_1 , x_2 , . . . , x_n )$.\footnotemark[1]}
 \label{fig:bilingual_diagram}
\end{figure}

We then leverage the obtained sentence embedding $M$ by using an attention-based decoder over its components similar to  \citet{cifka_bleu-VS-meaningRep_2018}. In Figure \ref{fig:bilingual_diagram} we present a diagram from the same work adapted to our formulation. Formally, we only need to compute equations \eqref{eq:alphas} and \eqref{eq:energy} using the columns of $M$ instead of the encoder states $h_i$.

 \footnotetext[1]{Adaptation from \citet{cifka_bleu-VS-meaningRep_2018}.}

Some ways of initializing the states of the decoder consist on using the last state of the encoder \cite{bahdanau_attention_2014}, the average of the encoder states \cite{sennrich_nematus_2017}, among others. Nevertheless, after comparing the results attained with them, we instead propose to use the average of the attention heads $m_i$'s, as follows  
\begin{align}
s_0 = tanh(W_{s_{0}} \overline{m}) \label{eq:s0}
\end{align}
where $W_{s_{0}} \in \mathbb{R}^{d_h \times d_h}$ and $\overline{m} = \frac{1}{r} \sum_{i=1}^r m_i$. This way, the decoder uses only information from the attention bridge. 

\textit{(ii) Language-specific encoders and decoders:}
To deal with additional language pairs, the model we propose incorporates a NN encoder for each input language and an attentive decoder for each output language. This adjusts the parameters of the attention bridge with multilingual information.

\begin{figure}[ht]
\flushright
\includegraphics[width= 1\hsize]{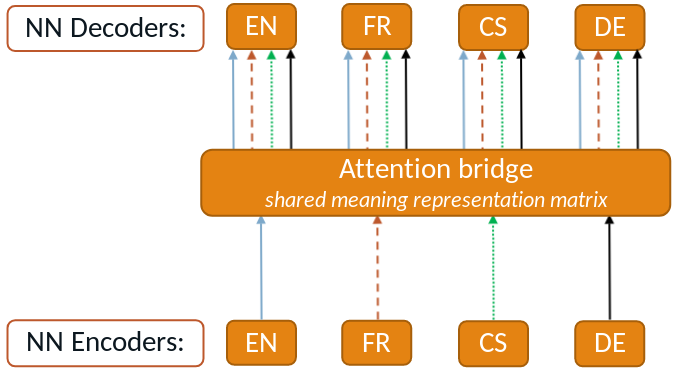} 
\caption{Diagram of the multiple encoders and decoders with and additional self-attention layer}
 \label{fig:multilingual_diagram}
\end{figure}

 Figure \ref{fig:multilingual_diagram} shows a basic diagram to illustrate the use of several encoders and decoders that are plugged in and out at every change of batch. During training it is important to shuffle the batches to avoid overfitting the attention bridge to one specific language pair; we have used a uniform distribution in this paper for this process. 

\textit{(iii) Penalization term:} The proposed attention bridge matrix $M$ from equation \eqref{eq:attention_bridge} could easily learn repetitive information for different attention heads. Since we want this representation to illustrate various components of a sentence, we introduce a penalty term in the model. Taking inspiration from recent work \cite{lin_self-attentive_2017,tao_utterance_2018}, we use the squared Frobenius norm of the symmetric matrix $(BB^T - I)$, with $B$ as defined in Eq. \eqref{eq:B}, as a redundancy measure, which is added to the loss function derived form Eq. \eqref{eq:loss}. Hence, our loss function becomes
\begin{align*}
\mathcal{L} = - \mbox{log}\left( p\left( X | Y \right) \right) +  \left\| BB^T - I \right\|^2_F
\end{align*}
By incorporating this term into the loss function we force matrix $BB^T$ to be similar to the identity matrix, that is, $\sum_j b_{ij}b_{ji} \approx 1$. Additionally, considering the fact that the rows of $B$ sum to 1, with entries in $[0,1]$ because it approximates a discrete probability distribution, it follows that the columns of $B$ will be forced to be approximately orthogonal, and hence we are penalizing redundancy.

\section{Experimental Setup}
\label{sec:experimental_setup}
To test the proposed architecture, we conducted four translation experiments. We used the multi30k dataset \cite{multi30k}, a multi-parallel dataset containing 29k image captions for training and 1k sentences for validation in four European languages; Czech (cs), German (de), French (fr) and English (en). We tested the trained model with the flickr 2016 test data of the same dataset and obtained BLEU scores using the sacreBLEU script \footnote[2]{with signature BLEU+case.lc+numrefs.1+smooth.exp+\\tok.13a+version.1.2.11} \cite{sacreBLEU}.

For each language, we used an encoder consisting of 2 stacked BiLSTMs of size $d_h = 512$, i.e., the hidden states for each direction are of size 256.
The neural interlingua layer we used has 10 attention heads, with a hidden dimension of 512 and $d_w =1024$, the dimensions of the linear transformations needed to compute matrix $B$ from Eq. (\ref{eq:B}).
The decoder consists of 2 stacked unidirectional LSTMs with hidden states of size 512. Further, the word embeddings we used have dimension $d_x = d_y = 512$, for both the encoder input and the decoder output. 

We used an SGD optimizer with a learning rate of 1.0 and batch size 64 and for each experiment we selected the best model on the development set.


We implemented our model on top of an OpenNMT-py \cite{opennmt} fork, which we make available for reproducibility purposes.\footnote[3]{https://github.com/Helsinki-NLP/OpenNMT-py/tree/neural-interlingua}

\section{Results}
\label{sec:results}

In the following, we present a number of experiments that demonstrate the capabilities of the attention bridge as an effective approach to multilingual MT. We first present our baselines, then we discuss many-to-one and one-to-many settings and, finally summarize our experiments on many-to-many translation models.

\subsection{Baselines}

The first experiment we conducted was to corroborate that the proposed architecture works properly 
by examining its performance in a bilingual setting. We expect that the models slightly drop in performance due to the fixed-size attention bridge that we introduce instead of directly mapping from source to target language via cross-lingual attention links. However, we want to see whether the architecture is robust enough to carry over the essential information needed for translation with the inclusion additional intermediate abstraction layer.

In Table~\ref{tab:bilingual_sanity-check} we present a comparison of our architecture in contrast with a strong bilingual baseline consisting of an architecture with the specifications described in section \ref{sec:experimental_setup}, without the components of our model. The table presents the scores obtained for each of the 12 bilingual models trained on each language pair. 

\begin{table}[ht]
  \small
  \begin{center}
    {\small
      \begin{tabular}{l  cccc}
        \toprule
        \multicolumn{5}{c}{\textsc{Bilingual}} \\
	    &	\textsc{en}	&	\textsc{de}	&	\textsc{cs}	&	\textsc{fr}	\\         \midrule
\textsc{en}	&	-	    &	36.78	&	28.00	&	55.96	\\
\textsc{de}	&	39.00	&	-	    &	23.44	&	38.22	\\
\textsc{cs}	&	35.89	&	28.98	&	-	    &	36.44	\\
\textsc{fr}	&	49.54	&	32.92	&	25.98	&	-	\\

        \midrule
        \midrule
        \multicolumn{5}{c}{\textsc{Bilingual + attention bridge}} \\
	    &	\textsc{en}	&	\textsc{de}	&	\textsc{cs}	&	\textsc{fr}	\\         \midrule
\textsc{en}	&	-   	&	35.85	&	27.10	&	53.03	\\
\textsc{de}	&	38.19	&	-   	&	23.97	&	37.40	\\
\textsc{cs}	&	36.41	&	27.28	&	-   	&	36.41	\\
\textsc{fr}	&	48.93	&	31.70	&	25.96	&	-	\\

        \bottomrule
      \end{tabular}}
  \end{center}
      \caption{\label{tab:bilingual_sanity-check} Baseline models - comparison of BLEU scores obtained with bilingual models. All models share specifications, apart from the proposed changes to include the attention bridge layer for the second part of the table.}
\end{table}

In this case we observe that the basic bilingual models without any attention bridge have a slightly better performance in almost every case. The biggest drop can be observed for English-French with a difference of over 2 BLEU points, but this case is exceptional. For all other languages pairs this difference lies within a range of less than 1 BLEU point.

This behaviour was expected from the fact that the information from the encoder has to be summarized in the 10 heads of the self-attention layer without (multilingual) information from other encoders to boost the states of this bridge. 
Nevertheless, these tests justify the validity of the architecture; namely, that the attention bridge does not cause a significant problem for the translation model in the bilingual case. We will use the results of both bilingual models with and without attention bridge as our baselines for the comparison to the multilingual models that we describe below.

\subsection{Many-To-One and One-To-Many Models}

The power of the attention bridge comes from its ability to share information across various language pairs. We now look at the effect including of additional languages during training on the translation performance of individual language pairs. We start by training models that include many-to-one and one-to-many settings with English as target and source, respectively. This setup makes it possible to study the ability of zero-shot translation, i.e. the translation between languages that have not been seen together in the training data. Consequently, looking at zero-shot translation, we can test the abstraction capabilities of the attention bridge.


For the first experiment, we trained a \{De,Fr,Cs\}$\leftrightarrow$En model using the many-to-one and one-to-many strategy as discussed above.
As depicted in Table \ref{tab:many2en}, this attempt 
already resulted in substantial improvements for the language pairs seen during training. The model exceeds both bilingual baselines from the previous section in all of these but French to English.
However, the model is completely incapable of performing zero-shot translations.
We believe that this inability of the model to generalize to unseen language pairs arises from the fact that every non-English encoder (or decoder) only learned to process information that was to be decoded into English (or encoded from English input). This finding is consistent with the results of \citet{lu_neural-interlingua_2018}. 

\begin{table}[ht]
  \small
  \begin{center}
    {\small
      \begin{tabular}{l  cccc}
        \toprule
        \multicolumn{5}{c}{\textsc{$\{$de,fr,cs$\} \leftrightarrow $ en}} \\
	&	\textsc{en}	&	\textsc{de}	&	\textsc{cs}	&	\textsc{fr}	\\         \midrule
\textsc{en}	&	-	    &	37.85	&	29.51	&	57.87	\\
\textsc{de}	&	39.39	&	-	    &	\cellcolor{blue!20} 0.35    &	\cellcolor{blue!20} 0.83	\\
\textsc{cs}	&	37.20	& \cellcolor{blue!20} 0.65	&	-   	&	\cellcolor{blue!20} 1.02	\\
\textsc{fr}	&	48.49	&	\cellcolor{blue!20} 0.60	&	\cellcolor{blue!20} 0.30	&	-	    \\

        \midrule
        \midrule
        \multicolumn{5}{c}{\textsc{$\{$de,fr,cs$\} \leftrightarrow $ en + monolingual}} \\
	&	\textsc{en}	&	\textsc{de}	&	\textsc{cs}	&	\textsc{fr}	\\         \midrule
\textsc{en}	&	-   	&	38.92	&	30.27	&	57.87 \\
\textsc{de}	&	40.17	&	-   	&	\cellcolor{blue!20} 19.50	&	\cellcolor{blue!20} 26.46 \\
\textsc{cs}	&	37.30	&	\cellcolor{blue!20} 22.13	&	-	    &	\cellcolor{blue!20} 22.80 \\
\textsc{fr}	&	50.41	&	\cellcolor{blue!20} 25.96	&	\cellcolor{blue!20} 20.09	&	-    \\

        \bottomrule
      \end{tabular}}
  \end{center}
  \caption{\label{tab:many2en} BLEU scores obtained for models trained on \{De,Fr,Cs\}$\leftrightarrow$En. Zero-shot translation (shaded cells) achieves noteworthy translation quality only when incorporating monolingual data during training.}
\end{table}

In order to address this problem, we incorporate monolingual data in training, that is, training $A \rightarrow A$ for each available language $A$ with identical copies of the input sentence as the target. In other words, no additional data was included during training, but we reincorporate examples from the same parallel training corpus used in all other experiments.  As a consequence, we see a remarkable increase in the BLEU scores, including a substantial boost for the language pairs not seen during training. 
In short, the monolingual data informs the model that other languages can be produced besides English, and that English is not the unique source language. 

Additionally, there is a positive effect on the seen language pairs, the cause of which is not immediately evident. One possibility may be that the shared layer acquires additional information that can be included in the abstraction process yet is not available to the other models.

\subsection{Many-to-Many Models}

To further examine the capabilities of the proposed architecture we conducted two experiments under a many-to-many scenario.

\begin{figure}[ht]
\flushleft
\hspace{-20pt}
\includegraphics[width= 1.08\columnwidth]{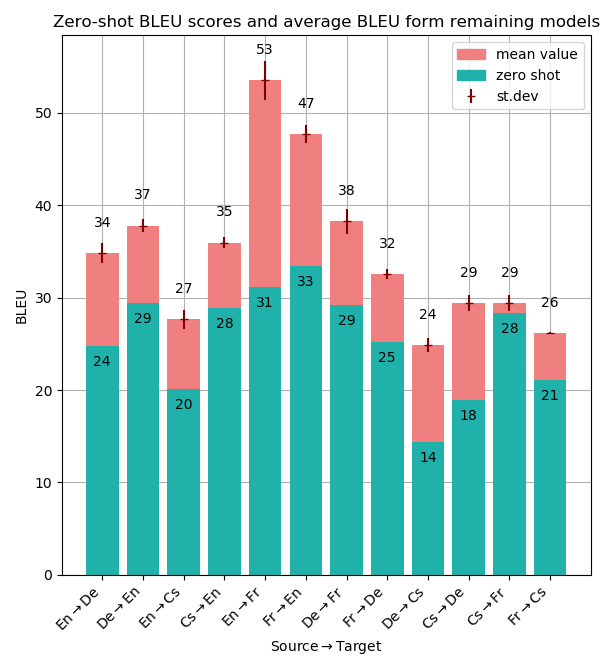} 
\caption{Zero-shot BLEU scores plot. The reported mean value and standard deviation were computed using the BLEU scores from the remaining 5 models.}
 \label{fig:BLEU_plots}
\end{figure}

First, we trained six different models where we included all but one of the available language pairs. We then tested our models while also performing bidirectional zero-shot translations for the unseen language pairs. Figure \ref{fig:BLEU_plots} summarizes these results, where we report the scores of the zero-shot translation for each source and target language. The figure shows as well the mean and standard deviation of the BLEU scores obtained by the remaining 5 models that did see the respective source and target languages during training. We observe that the zero-shot translation scores are generally better than ones from the previous \{De,Fr,Cs\}$\leftrightarrow$En model with monolingual data, even though in this set of experiments we did not include monolingual data.



Finally, we also tested the architecture in a many-to-many setting with all language pairs included. 

Table~\ref{tab:many2many} summarizes the results of our experiments. As in the previous case, we compare settings that include monolingual data with their counterparts that do not include it. 

On a first note, the inclusion of language pairs results in an improved performance when compared to the bilingual baselines, as well as the many-to-one and one-to-many cases. The only exception being the \textsc{En}$\rightarrow$\textsc{Fr} task. 
Moreover, the addition of monolingual data during training leads to even higher scores, producing the overall best model. The absolute improvements in BLEU range from 1.40 to 4.43 compared to the standard bilingual model. 

% ...


\begin{table}[ht]
  \small
  \begin{center}
    {\small
      \begin{tabular}{l  cccc}
        \toprule
        \multicolumn{5}{c}{\textsc{m-2-m}} \\
	&	\textsc{en}	&	\textsc{de}	&	\textsc{cs}	&	\textsc{fr}	\\         \midrule
\textsc{en}	&	-	    &	37.70	&	29.67	&	55.78	\\
\textsc{de}	&	40.68	&	-	    &	26.78   &	41.07	\\
\textsc{cs}	&	38.42	&	31.07	&	-   	&	40.27	\\
\textsc{fr}	&	49.92	&	34.63   &	26.92	&	-	    \\

        \midrule
        \midrule
        \multicolumn{5}{c}{\textsc{m-2-m + monolingual}} \\
	&	\textsc{en}	&	\textsc{de}	&	\textsc{cs}	&	\textsc{fr}	\\         \midrule
\textsc{en}	&	-   	&	38.48	&	30.47	&	57.35 \\
\textsc{de}	&	41.82	&	-   	&	26.90	&	41.49 \\
\textsc{cs}	&	39.58	&	31.51	&	-	    &	40.87 \\
\textsc{fr}	&	50.94	&	35.25	&	28.80	&	-    \\

        \bottomrule
      \end{tabular}}
  \end{center}
  \caption{\label{tab:many2many} The multilingual model also gets a boost when incorporating monolingual data during training.}
\end{table}

We performed additional tests using Transformer-based encoders. These results were, however, unsatisfactory, with BLEU scores well below their RNN counterpars for all language pairs. The inability of these models to obtain good translation quality might arise from the fact that the multi30K dataset is rather small, making the Transformer prone to overfitting due to its large amount of parameters.


\section{Conclusion}
\label{sec:conclusion}
In this work we propose a multilingual NMT architecture with three modifications to the common attentive encoder-decoder architecture. By introducing language-specific encoders and decoders, a shared language-independent attention bridge and a penalization term that forces this layer to attend different semantic structures of the input sentence, we accomplish to successfully develop a strong multilingual translation system that efficiently incorporates transfer learning and can also tackle the task of learning multilingual sentence representations. The attention bridge consists of a self-attention layer shared among all languages that can be regarded as a neural interlingua due to its capacity of abstracting and handling encoded multilingual information.

Furthermore, we performed four different experiments to demonstrate the capabilities of the attention bridge architecture as an effective approach to multilingual  MT. The results obtained consistently outperform a strong bilingual model, which suggests that the attention bridge layer has the ability to efficiently share parameters in a multilingual setting. The inclusion of monolingual data during training resulted in boosted scores for all cases.

Future work stemming from these results includes downstream-testing the sentence meaning representations produced by the shared attention bridge to verify its generalization capabilities. In addition, training models on larger datasets as well as reporting the effects of using non-multiparallel datasets would expand the scope of this work.

\section*{Acknowledgments}

  \begin{wrapfigure}{r}{0.3\hsize}
  \flushleft
    \vspace{-20pt}
    \includegraphics[width=.7\hsize]{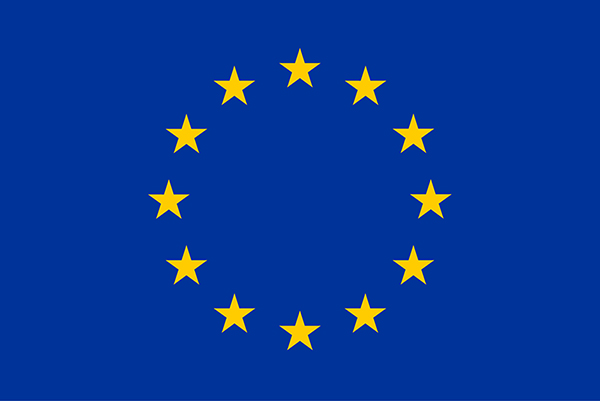}
    \vspace{-10pt} \hspace{-10pt}
\end{wrapfigure}
  
 This project has received funding from the European Research Council (ERC) under the European Union’s Horizon 2020 research and innovation programme (grant agreement No 771113)

We thank the participants that contributed to the project we lead during the 13th MT Marathon in Prague. We are particularly grateful with Chris Hokamp, whose help was crucial during that time. Finally, We would also like to acknowledge NVIDIA and their GPU grant.
\bibliography{naaclhlt2019}
\bibliographystyle{acl_natbib}

\end{document}